\documentclass{article}
\usepackage{spconf}
\usepackage[cmex10]{amsmath}
\usepackage{amsthm}
\usepackage{amssymb}
\usepackage{mathrsfs}
\usepackage{graphicx}
\usepackage{float}
\usepackage{array}
\usepackage{epstopdf}
\usepackage{multirow}

\title{a\MakeLowercase{n} e\MakeLowercase{xperimental} s\MakeLowercase{tudy} \MakeLowercase{of} D\MakeLowercase{eep} c\MakeLowercase{onvolutional} f\MakeLowercase{eatures} f\MakeLowercase{or} 
 i\MakeLowercase{ris} r\MakeLowercase{ecognition}}

\name{Shervin Minaee$^*$, Amirali Abdolrashidi$^{\dagger}$ and Yao Wang$^*$}
\address{$^*$Electrical Engineering Department, New York University,
\\ $^{\dagger}$Computer Science and  Engineering Department, University of California at Riverside}

\allowdisplaybreaks[1]	
\begin{document}

\maketitle

\begin{abstract}
Iris is one of the popular biometrics that is widely used for identity authentication. 
Different features have been used to perform iris recognition in the past. Most of them are based on hand-crafted features designed by biometrics experts. 
Due to tremendous success of deep learning in computer vision problems, there has been a lot of interest in applying features learned by convolutional neural networks on general image recognition to other tasks such as segmentation, face recognition, and object detection.
In this paper, we have investigated the application of deep features extracted from VGG-Net for iris recognition.
The proposed scheme has been tested on two well-known iris databases, and has shown promising results with the best accuracy rate of 99.4\%, which outperforms the previous best result.
\end{abstract}

\section{Introduction}
\label{sec:intro}
To make an application personal or more secure, we need to be able to distinguish a person from everyone else. 
Biometric features are a popular way for authentication, which cannot be imitated by any other than the desired person himself.  
Many works revolve around identification and verification of biometric data including, but not limited to, faces, fingerprints, iris patterns, and palmprints \cite{Face}-\cite{palm_wave}.
Iris recognition systems are widely used for security applications, since they contain a rich set of features and do not change significantly over time.

There have been many algorithms for iris recognition in the past. 
One of the early algorithms was developed by John Daugman that used 2D Gabor wavelet transform \cite{Daugman}. 
In a more recent work, Kumar \cite{database} proposed to use a combination of features based on Haar wavelet, DCT and FFT to achieve high accuracy. In \cite{Farouk}, Farouk proposed a scheme which uses elastic graph matching and Gabor wavelet. 
Each iris is represented as a labeled graph and a similarity function is defined to compare the two graphs. 
In \cite{Belcher}, Belcher used region-based SIFT descriptor for iris recognition and achieved a relatively good performance. Pillai \cite{Pillai} proposed a unified framework based on random projections and sparse representations to achieve robust and accurate iris matching. 
In a more recent work \cite{myscat}, Minaee proposed an algorithm based on textural and scattering transform features, which achieved a high accuracy rate compared to the previous schemes. This approach also extracts features from a multi-layer representation, but with predefined filters. 
The reader is referred to \cite{Iris} for a comprehensive survey of iris recognition.

In most of iris recognition works, first the iris region is segmented and then it is mapped to a rectangular region in polar coordinates. After that, the features are extracted from this region.
Most of these features are hand-crafted to work well for a given type of data or biometrics.
The main issue with the traditional approaches is that they require a lot of pre-processing and parameter tuning to work reasonably well for a given dataset, and their efficiency on other biometrics or even a different dataset of the same biometric is not guaranteed.
To overcome this issue, there has been a lot of effort in the past few years to learn some general features which can be transferred across many tasks. 
Along this direction, deep neural networks have achieved state-of-the-art results on various datasets, most notably AlexNet \cite{AlexNet}, which is trained on ImageNet competition (which contains around 1.2 million hand-labeled images for 1000 categories) \cite{ImageNet}. 
In the deep learning framework, the images are fed as the input of the multi-layer neural network and the network discovers the best way to combine the pixels for maximizing the recognition accuracy.
In \cite{shelf}, through an experimental study, it is shown that the features learned by training a deep network on image recognition, can be transferred to other tasks and datasets and achieve a remarkable performance.
Since then, features from different networks, such as Alex-Net, ZF-Net, VGG-Net and Res-Net \cite{AlexNet}-\cite{resnet} have been used for various tasks, such as texture synthesis, object detection and, image segmentation.

In this work, we explored the application of deep features extracted from VGG-Net for iris recognition task.
We treat the trained model as a feature extraction engine and use it for feature extraction from iris images, without any fine-tuning to see if the general features are applicable for biometric recognition. 
Then PCA is applied to reduce feature dimensionality and then multi-class SVM is used to perform recognition.
We have provided extensive experimental study on two well-known iris datasets, CASIA 1000 Iris dataset \cite{CASIA} and IIT iris dataset \cite{IIT}.
Four sample iris images from CASIA-1000 dataset are shown in Figure 1.
\begin{figure}[1 h]
\begin{center}
    \includegraphics [scale=0.3] {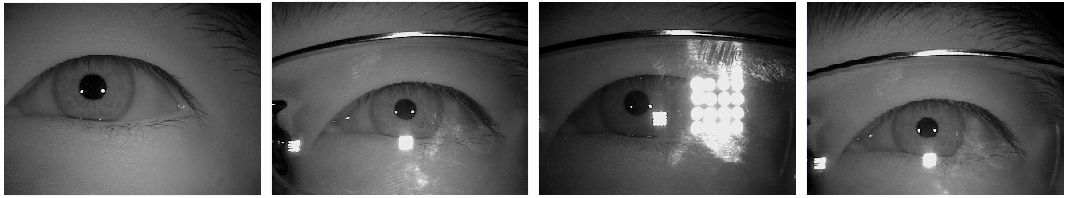}
\end{center}
  \vspace{-0.3cm}
  \caption{Sample images from CASIA-1000 dataset}
\end{figure}

It is worth mentioning that in our framework, we have skipped the segmentation step to see the robustness of these features to intra-class variation, and despite the existence of many variations in CASIA-1000 dataset, this algorithm has achieved a very high-accuracy rate.

It is quite interesting that, although VGG-Net is trained to classify objects from different categories, the CNN feature from this network works reasonably well for iris recognition which is to classify iris images of different subjects (i.e. all images belong to the same object category, here iris).

The rest of the paper is organized as follows. Section \ref{SectionII} describes the deep features that are used in this work, and gives a quick overview of VGG-Net architecture. Section \ref{SectionIII} provides a brief description of the dimensionality reduction and classification scheme used in this work. 
The experimental results and comparisons with other works are presented in Section \ref{SectionIV}, and the paper is concluded in Section \ref{SectionV}.

\section{Features}
\label{SectionII}
Extracting good features and image descriptors is a very important step in many computer vision and object recognition algorithms. 
Many features have been designed during the past years, which provide good representation for many image categories, to name a few,  scale-invariant feature transform (SIFT), histogram of oriented gradient (HOG), and bag of words (BoW) \cite{sift}-\cite{BoW}.
These features are mostly designed by computer vision experts, and are usually referred to as hand-crafted features.
Different applications, such as medical image analysis \cite{omid1}-\cite{omid2}, may use a very different set of hand-crafted features.
Recently, feature-learning algorithms and multi-layer representation have drawn a lot of attention, most notably convolutional neural networks \cite{lecun}, where the image is fed directly as the input to the deep neural network and the algorithm itself finds the best set of features from the image.
Deep architecture based on convolutional neural networks achieved state-of-the-art results in many computer vision benchmarks during past few years, including AlexNet, ZF-Net, GoogLeNet, VGG-Net and ResNet, which achieved promising results in ImageNet competition in 2012 and 2015. 
The features learned from these networks are shown to be well transferred to many other datasets and tasks \cite{shelf}, which means the trained network on ImageNet images is used to extract features from other dataset and then a classifier is trained on top of these features to perform recognition in other datasets, such as Caltech101 or Caltech256.
They have also been explored for a wide range of other computer vision tasks, such as segmentation, object tracking, texture synthesis, colorization, and super-resolution \cite{seg}-\cite{SR}.

It would be interesting to analyze the application of deep features learned by training a ConvNet, for the iris recognition problem, which is different from object recognition in the sense that the goal is not to distinguish one type of object from another, but to distinguish images of the same kind belonging to different people.
In this work, the application of deep features, extracted from VGG-Net, is explored for iris recognition, which to the best of our knowledge has not been studied yet. Here we give a quick overview of VGG-Net.

\subsection{VGG-Net Architecture}
VGG-Net was the runner-up in ILSVRC 2014 (ImageNet Large Scale Visual Recognition Competition), proposed by Karen Simonyan and Andrew Zisserman \cite{VGG}. 
Figure 2 shows the overall architecture of VGG-Net.
Its showed that the depth of the network is a critical component for good performance.
Their final best network contains 16 CONV/FC layers, and 5 pooling layers.
In total, it has around 138 million parameters (which most of it is in the first FC layer that contains 102M weights).
One of the nice thing about VGG-Net is its extremely homogeneous architecture, which only performs 3x3 convolutions with stride 1 and pad 1, and 2x2 pooling (with no padding) from the beginning to the end \cite{stan}.

We refer the reader to \cite{VGG} and \cite{stan} for further details about the architecture and number of parameters of each layer of this network. 
We extract features from different layers of this network and evaluate their performance for iris recognition.
\begin{figure}[2 h]
\begin{center}
    \includegraphics [scale=0.30] {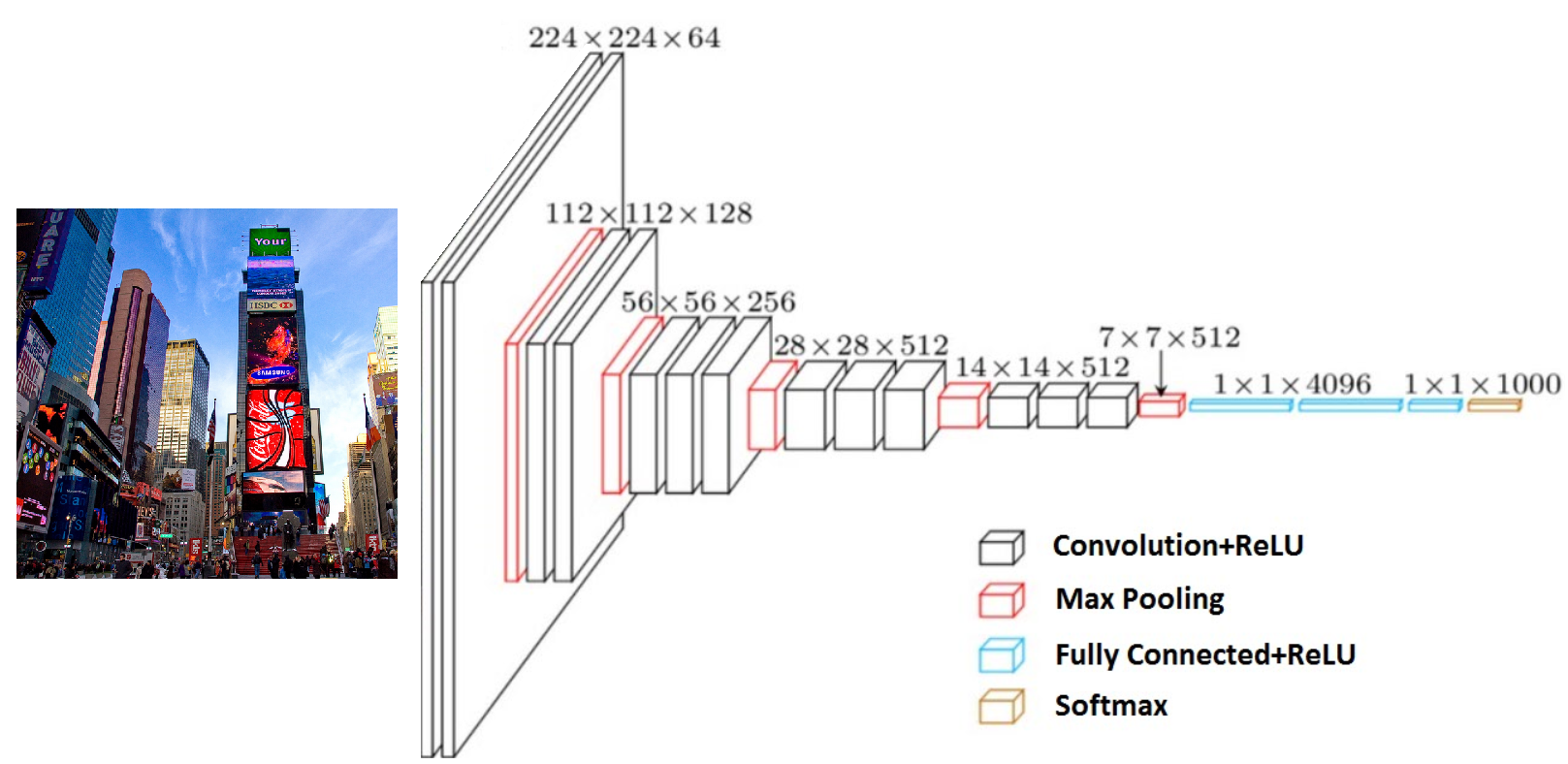}
\end{center}
\vspace{-0.2cm}
  \caption{Architecture of VGG16 \cite{vgg_img}}
\end{figure}

We further perform PCA \cite{PCA} on the deep features to reduce their dimensionality, and evaluate the performance of the proposed algorithm for different feature dimensions.

\subsection{Alternative Hierarchical Representation With Predefined Filters}
We would like to mention that there are also other kinds of deep architectures, which do not learn the filters and use predefined weights. Scattering convolutional network is one such network that decomposes the signal into multiple layers of different scales and orientations \cite{mallat}.
Then some statistical features are extracted from each transform image, and features from all transformed images are concatenated to form the overall feature representation. 
The scattering network has also been used for biometric recognition in the past \cite{palm_sh}- \cite{face_sh}.
To illustrate how this network works, the output of the second layer of scattering network for a sample iris image is shown in Figure 3. As it can be seen, each image contains the edge information along some orientation in some scale. However, since the filters are not learned and are fixed across different tasks, this architecture may not be the optimum one for some of the visual applications. Yet, they could be very useful for the cases the amount of training data is very limited and is not enough to train and convolutional neural net.
\begin{figure}[2 h]
\begin{center}
    \includegraphics [scale=0.3] {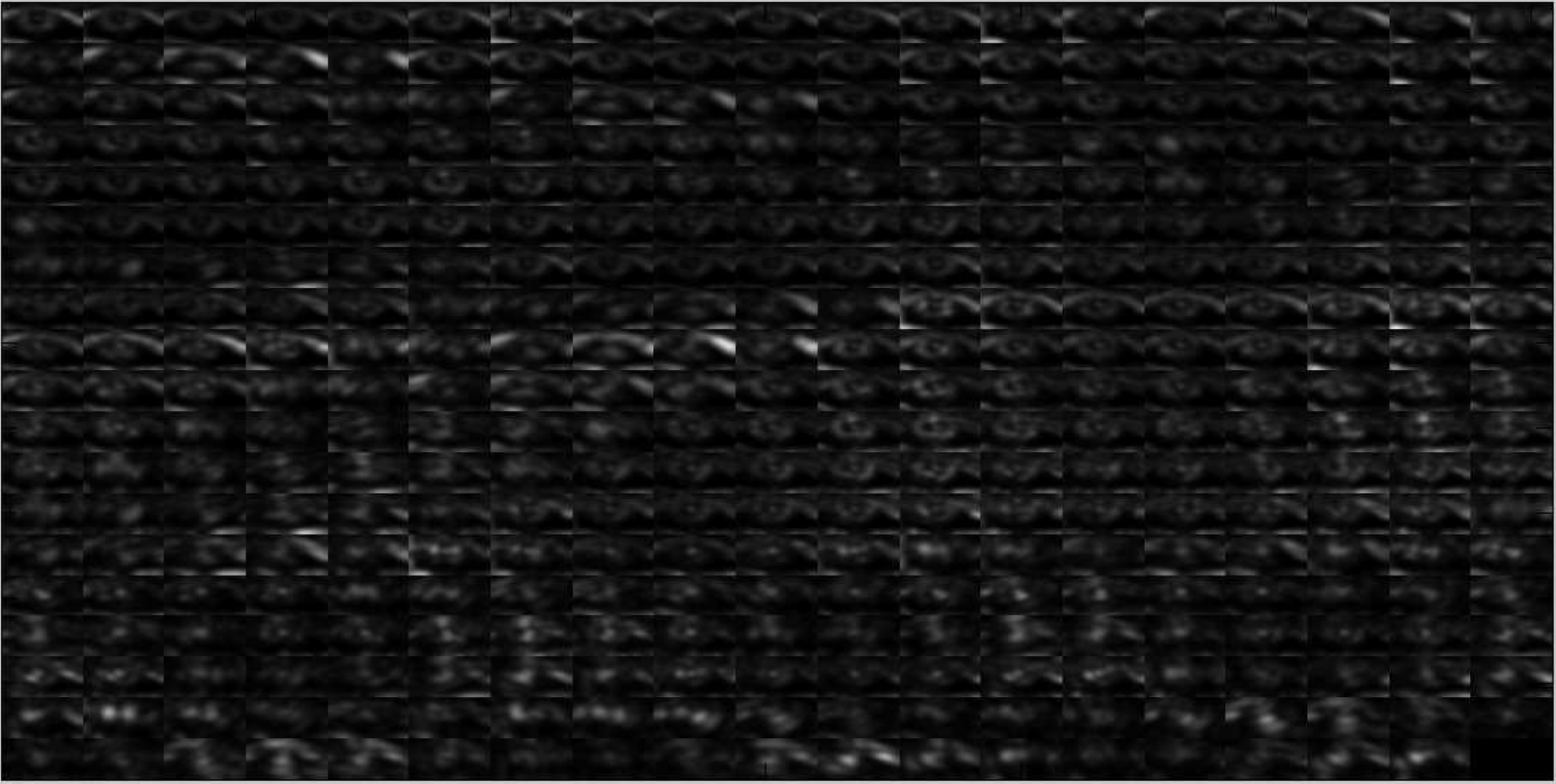}
\end{center}
\vspace{-0.2cm}
  \caption{The images of the second layer of scattering network}
\end{figure}
\vspace{-0.5cm}

\section{Recognition Algorithm}
\label{SectionIII}
After feature extraction, a classifier is needed to find the corresponding label for each test image.
There are various classifiers which can be used for this task, including softmax regression (or multinomial logistic regression), support vector machine \cite{SVM} and neural network \cite{NN}. In this work, multi-class SVM has been used which is quite popular for image classification.
A brief overview of SVM for binary classification is presented here. For further details and extensions to multi-class settings and nonlinear version, we refer the reader to \cite{multi_SVM} and \cite{kSVM}.
Let us assume we want to separate the set of training data $(x_1,y_1)$, $(x_2,y_2)$, ..., $(x_n,y_n)$ into two classes where 
$x_i \in \mathbf{R}^d $ is the feature vector and $y_i \in \{-1,+1\}$ is the class label. 
Assuming the two classes are linearly separable, we can separate them with a hyperplane $w.x+b=0$. Among all possible choices of hyperplanes SVM finds the one with the maximum margin. 
One can show that the maximum margin hyperplane can be found by the following optimization problem:
\begin{equation}
\begin{aligned}
& \underset{w,b}{\text{minimize}}
& & \frac{1}{2} ||w||^2 \\
& \text{subject to}
& & y_i(w.x_i+b) \geq 1, \; i = 1, \ldots, n.
\end{aligned}
\end{equation}
The above optimization is convex, and therefore it can be solved with convex optimization techniques.
One popular way is to solve the dual problem by introducing Lagrange multipliers $\alpha_i$, which results in a classifier as $f(x)= sign(\sum_{i=1}^{n} \alpha_i y_i w.x+b)$,
where $\alpha_i$'s and $b$ are calculated by the SVM learning algorithm. 
There is also a soft-margin version of SVM which allows for mislabeled examples. 
To perform multi-class SVM for a set of data with $M$ classes, we can train $M$ binary classifiers which can discriminate each class against all other classes, and to choose the class which classifies the test sample with the largest margin (one-vs-all). 
In another approach, we can train  a set of $M \choose 2$ binary classifiers, each of which separates one class from another one and then choose the class that is selected by the majority of classifiers. 
There are also other approaches to perform SVM for multi-class classification.

\section{Experimental results and analysis}
\label{SectionIV}
In this section, a detailed description of our experimental results is provided.
Before presenting the results, let us describe the parameter values of our algorithm.
For each image, we extract features from different layers of VGG-Net, from fc6 and some of the previous layers. The output of fc6 layer is a 4096 dimensional vector, but in the layers before that, we have 256/512 filter outputs of different sizes. 
We take the average of each filter output and form a feature vector from each layer to evaluate the performance of features extracted from different layers. 
For SVM, we have used the LIBSVM library \cite{libsvm}, and linear kernel is used with the penalty cost $C=1$.

We have tested our algorithm on two iris databases, CASIA-Iris-1000 and IIT Delhi.
CASIA-Iris-1000 contains 20,000 iris images from 1,000 subjects, which were collected using an IKEMB-100 camera \cite{CASIA}. The main sources of intra-class variations in CASIA-Iris-1000 are eyeglasses and specular reflections.
The IIT Delhi database contains 2240 iris images captured from 224 different people.
The resolution of these images is 320x240 pixels \cite{IIT}. We resize all images to 224x224 to be suitable for VGG-Net input.

For each person, around half of the images are used for training and the rest for testing.
We first evaluate the recognition accuracy of features from the fc6 layer. First PCA is applied to all features and the recognition accuracy for different number of PCA features evaluated. 
Figures 4 and 5 show the recognition accuracy using different numbers of PCA features for IIT Delhi and CASIA datasets respectively.
Interestingly, even by using few PCA features, we are able to get a very high accuracy rate. 
As it can be seen, using 100 PCA features results in an accuracy rate above 98\% for IIT database, which only increases around 1\% by using more PCA features.

\begin{figure}[3 h]
\begin{center}
    \includegraphics [scale=0.46] {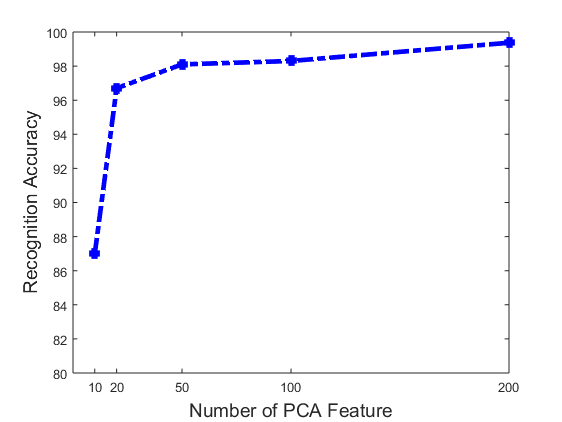}
    \vspace{-0.3cm}
\end{center}
  \caption{Recognition accuracy for different numbers of PCA features on IIT Delhi Iris Database}
\end{figure}

\begin{figure}[3 h]
\begin{center}
    \includegraphics [scale=0.46] {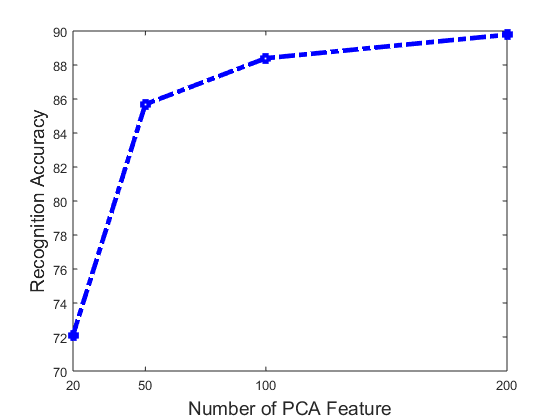}
    \vspace{-0.3cm}
\end{center}
  \caption{Recognition accuracy for different numbers of PCA features on CASIA-1000 Iris Database}
\end{figure}

\begin{figure}[3 h]
\begin{center}
    \includegraphics [scale=0.46] {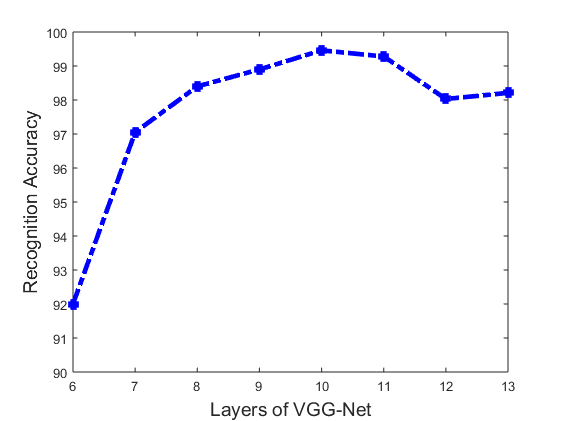}
    \vspace{-0.3cm}
\end{center}
  \caption{Recognition accuracy for features of different layers of VGG-Net on IIT Delhi Iris Database}
\end{figure}

In another experiment, we evaluated the performance of deep features extracted from different layers of VGG-Net.
To have a fair comparison, we restrict the number of features from each layer to be 256 (by taking first 256 PCA, if that layer has more than 256 output filters). 
Figure 6 shows the recognition accuracy achieved by features from different layers of VGG-Net on IIT Delhi database.

As it can be seen, by extracting features from any layer after the 7th, an accuracy rate of above 98\% can be achieved.
The recognition accuracy achieves its peak by using features from the 10th layer and then drops. One possible reason could be that by going into the higher layers of the deep network, they start to capture more abstract and high-level information which does not discriminate much between different iris patterns, whereas the mid-level features in the previous layers have more discriminating power for same-class recognition.

We have also evaluated the robustness of the proposed scheme to the number of training samples used for training.
We varied the number of training samples for each person from 1 to 5 (out of 10 samples) and found the recognition accuracy.
Figure 7 shows the accuracy of this algorithm for different training samples for IIT database.
As it can be seen, the recognition accuracy by using 3 samples out of 10, has a large gain over using 1 or 2 samples, and it remains relatively constant by increasing the number of training samples.

\begin{figure}[5 h]
\begin{center}
    \includegraphics [scale=0.46] {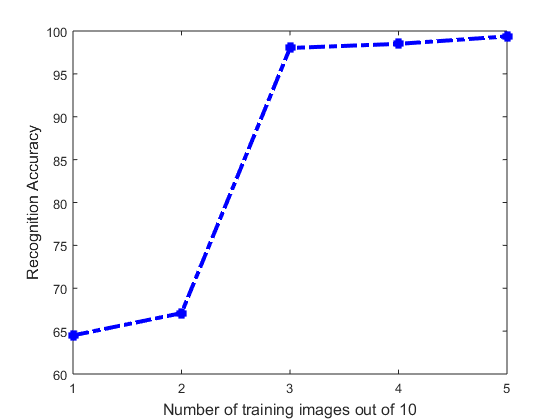}
    \vspace{-0.3cm}
\end{center}
  \caption{Recognition rate versus number of training samples}
\end{figure}

Table 1 provides a comparison of the performance of the proposed scheme and those of other recent algorithms on IIT database.
The scattering transform scheme [10] also uses a multi-layer representation and achieves a very high accuracy rate. 
By using the deep features, we were able to achieve the highest accuracy rate on this dataset.
This is mainly due to the richness of deep features which are able to capture many of the information lost in hand-crafted features, providing a very high discriminating power.
One main advantage of this scheme is that, it does not require segmentation of iris from eye images although the segmentation could improve the results for some difficult cases.

\begin{table} [h]
\centering
  \caption{Comparison of the proposed algorithm with the previous algorithms on IIT database}
  \centering
\begin{tabular}{|m{6.2cm}|m{1.6cm}|}
\hline
\ \ \ \ \ \ \ \ \ \ \ \ \ \ \ \ \ \ \ \ \ \ \ \ \ Method &  Recognition \ \hspace{4cm} Rate\\
\hline
Haar Wavelet \cite{database} & \ \ \ \ \  96.6\% \\
\hline
Log Gabor Filter by Kumar \cite{database} & \ \ \ \ \ 97.2\% \\
\hline
Fusion \cite{database} & \ \ \ \ \  97.4\% \\
\hline
Elastic Graph Matching \cite{Farouk} & \ \ \ \ \ 98\% \\
\hline
Texture+Scattering Features \cite{myscat} & \ \ \ \ \ 99.2\% \\
\hline
\textbf{Proposed scheme} & \ \ \ \ \  \textbf{99.4\%} \\
\hline
\end{tabular}
\label{TblComp}
\end{table}

The experiments are performed using MATLAB 2015 on a laptop with Core i5 CPU running at 2.2GHz. 
For deep learning features, the MatConvNet package is used to extract the features from VGG-Net \cite{matconv}.

\section{Conclusion}
\label{SectionV}
In this work, we evaluated the application of deep features, extracted from VGG-Net, followed by a simple classification algorithm for  the problem of iris recognition. 
Deep features have been in the center of attention during these years and are being used for many different applications.
Although the original convolutional network used in this work is trained for a very different task (object recognition), it is shown that the features can be well transferred to biometric recognition.
This algorithm is tested on two well-known datasets and achieved promising results, which outperforms the previous best results  on one of the databases. 
We want to note that most of the previous algorithms for iris recognition involve a lot of pre-processing and parameter tuning, but in our framework, no pre-processing and architecture optimization is performed. 
This result could be further improved by training a deep network specifically for iris recognition, which is left for future research.

\section*{Acknowledgments}
The authors would like to thank Vadaldi's group at Oxford University for providing the software implementation of convolutional neural network, and also CSIE group at NTU for providing the LIBSVM package. 
We would also like to thank CASIA center and IIT Delhi for providing the iris databases.

\end{document}